\documentclass[conference]{IEEEtran}      

\IEEEoverridecommandlockouts                              

\usepackage{graphicx} 
\usepackage{tensor}
\usepackage[font=footnotesize]{caption}
\usepackage{cite}
\usepackage[font=footnotesize]{subcaption}
\usepackage{booktabs}
\usepackage{times} 
\usepackage{bm}
\usepackage{amsmath,amssymb,mathrsfs,mathdesign,upgreek,dsfont,gensymb}
\usepackage{extarrows}
\usepackage[ruled,vlined]{algorithm2e}
\usepackage{verbatim}
\usepackage{diagbox}
\usepackage[hidelinks]{hyperref}
\hypersetup{
   linkcolor=blue,
   breaklinks=true,  
   colorlinks=true, 
   citecolor=blue,
   urlcolor=blue
}
\usepackage{cleveref}
\usepackage{soul}
\usepackage[multiple]{footmisc}
\usepackage{multirow}
\usepackage{threeparttable}

\usepackage{enumitem}
\usepackage{color}
\usepackage{xspace}
\usepackage{float}

\graphicspath{{./figures/}}

\DeclareCaptionSubType*[Alph]{table}
\DeclareCaptionLabelFormat{mystyle}{Table~\bothIfFirst{#1}{ ̃}#2}

\newcommand{\R}{\mathbb{R}}

\DeclareMathOperator*{\argmin}{arg\!\min}

\newcommand*\diff{\mathop{}\!\mathrm{d}}

\newcommand \drafto{DRAFTO\xspace}
\newcommand \figref{Figure \ref}
\newcommand \tabref{Table~\ref}

\newcommand \algref{Algorithm~\ref}

\newtheorem{thm}{Theorem}


\title{\textbf{\drafto}: \textbf{D}ecoupled \textbf{R}educed-space and \textbf{A}daptive \textbf{F}easibility-repair \\ \textbf{T}rajectory \textbf{O}ptimization for Robotic Manipulators}

\author{
Yichang Feng$^{1}$, Xiao Liang$^{2,*}$, and Minghui Zheng$^{1,*}$
\thanks{$^{1}$Yichang Feng and Minghui Zheng are with the J. Mike Walker '66 Department of Mechanical Engineering, Texas A\&M University, College Station, TX 77843, USA (\tt\footnotesize e-mail: yichangfeng@tamu.edu; mhzheng@tamu.edu).}
\thanks{$^{2}$Xiao Liang is with the Zachry Department of Civil and Environmental
Engineering, Texas A\&M University, College Station, TX 77843 USA (\tt\footnotesize e-mail:
xliang@tamu.edu).}
\thanks{$^*$ Corresponding Authors.}
\thanks{This work was supported by the USA National Science Foundation under Grant No. 2527316 and No. 2422826. The source code for this work will be made available in a public repository upon acceptance of the manuscript.}
}

\begin{document}

\bstctlcite{IEEE:BSTcontrol}

\maketitle
\thispagestyle{empty}
\pagestyle{empty}

\begin{abstract}

This paper introduces a new algorithm for trajectory optimization, Decoupled Reduced-space and Adaptive Feasibility-repair Trajectory Optimization (\drafto). It first constructs a constrained objective that accounts for smoothness, safety, joint limits, and task requirements. Then, it optimizes the coefficients, which are the coordinates of a set of basis functions for trajectory parameterization. To reduce the number of repeated constrained optimizations while handling joint-limit feasibility, the optimization is decoupled into a reduced-space Gauss-Newton (GN) descent for the main iterations and constrained quadratic programming for initialization and terminal feasibility repair. The two-phase acceptance rule with a non-monotone policy is applied to the GN model, which uses a hinge-squared penalty for inequality constraints, to ensure globalizability. The results of our benchmark tests against optimization-based planners, such as CHOMP, TrajOpt, GPMP2, and FACTO, and sampling-based planners, such as RRT-Connect, RRT*, and PRM, validate the high efficiency and reliability across diverse scenarios and tasks. The experiment involving grabbing an object from a drawer further demonstrates the potential for implementation in complex manipulation tasks. The supplemental video is available at \href{https://youtu.be/XisFI37YyTQ}{this link}.

\end{abstract}

\section{Introduction} 

With the rapid development of industrial intelligence and smart home technologies, the environment in which robots operate has become more complex and dynamic, and the tasks they face have become more diverse. This has placed higher demands on the reliability, efficiency, and flexibility of the underlying motion-planning algorithms for robots. 

Traditional motion planning algorithms are mainly divided into two categories: optimization-based planners and sampling-based planners. 
Optimization-based planners, such as CHOMP \cite{zucker_chomp_2013}, TrajOpt \cite{schulman_motion_2014}, and GPMP \cite{mukadam_continuous-time_2018}, are often efficient, and the feasible solutions are usually smoother. However, due to the non-convexity of the objective function, their solution process is prone to getting stuck in an infeasible local minimum. 
In contrast, sampling-based planners, such as RRT-Connect (RRT-C) \cite{kuffner_rrt-connect_2000}, RRT* \cite{karaman_sampling-based_2011}, and PRM \cite{kavraki_probabilistic_1996}, achieve probabilistic completeness. However, their uniform sampling and collision-detection processes significantly reduce planning efficiency. Moreover, most trajectories generated by sampling-based planners are too jerky to execute. Therefore, they often require post-processing. 
With the rise of AI, MPNet \cite{qureshi_motion_2019} and M$\pi$Net \cite{fishman_motion_2022} can effectively infer the feasible samples using neural networks. However, in addition to the significant computational resources required for network training, their generalization is affected by the training data and methods, and the trained models are inefficient on CPUs, which hinders the planning-execution workflow based on the RT-kernel. Also, unlike the Lagrangian multiplier-based CHOMP and IMACS \cite{kingston_exploring_2019} frameworks for sampling-based planners, most learning-based planners cannot plan task-constrained motion.  

\begin{figure}[!htp]
    \centering
    \includegraphics[width=.95\linewidth]{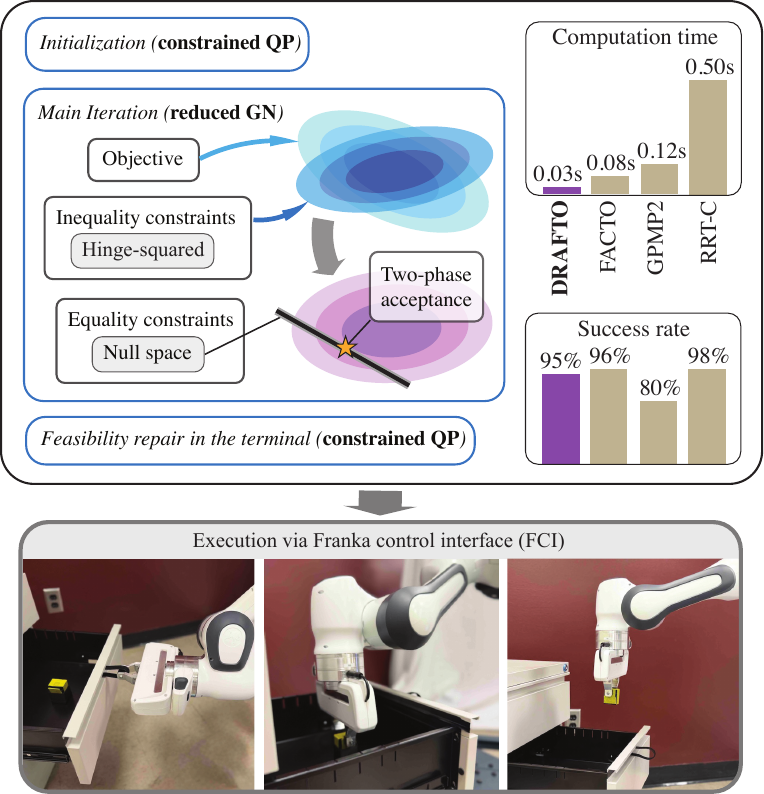}
    \caption{
    Overview of \drafto with decoupled optimization processes: (i) constrained-QP for initialization and terminal feasibility repair; (ii) reduced-space Gauss–Newton iterations with null-space equality handling and hinge-squared inequality penalties.  Comparison against FACTO, GPMP2, and RRT-Connect demonstrates high computational efficiency and success rate; a real-world drawer-grabbing execution via FCI is shown below. 
    }
    \label{fig:drafto}
\end{figure}

\drafto, decoupled reduced-space and adaptive feasibility-repair trajectory optimization shown in \figref{fig:drafto}, is proposed to address the above issues. It adopts the function-space framework of FACTO \cite{feng_facto_2026} to reduce the number of variables and the non-convexity arising from the time-continuous collision. Under this framework, the optimized trajectory with scaled execution time can be directly implemented on a Franka Research 3 (FR3) robot.   
To address the high computational cost of the massive inequality constraints, mainly arising from trajectory-wide joint limits, the optimization procedure is decoupled into reduced-space Gaussian-Newton (GN) steps and constrained quadratic programming (QP) steps. The constrained QP is primarily used for trajectory initialization, given the initial and goal waypoints, and for feasibility repair in the termination stage. The reduced-GN dominates the main iterations and handles equality and inequality constraints using the null-space and hinge-squared penalty methods, respectively. Furthermore, an adaptive regularization method is introduced to ensure that GN steps remain well-conditioned. The two-phase acceptance rule is applied to ensure the globalizability of \drafto. 

\paragraph*{Our contributions can be summarized as follows}
\begin{itemize}[itemindent=2\parindent, leftmargin=0pt] 
  \item A \textbf{decoupled framework} is proposed for the constrained trajectory optimization. It separates the penalty-based GN and constrained QP steps, significantly reducing the computational cost arising from the large number of inequality constraints. 
  \item  The\textbf{ two-phase acceptance rule} promotes globalization. The adaptive adjustment of the quasi-Hessian regulator improves the explorability in phase I. The non-monotone acceptance rule guarantees a stable convergence in phase II. 
  \item The \textbf{configurable checkpointing} works with the \textbf{null-space} method. It ensures the trajectory’s feasibility with respect to smoothness, safety, joint limits, and task constraints.
\end{itemize}

The benchmark results validate \drafto's high reliability and efficiency compared with optimization-based planners (CHOMP, TrajOpt, GPMP2, and FACTO) and sampling-based planners (RRT-Connect, RRT*, PRM) across more than 1,000 planning tasks in both the single-arm and dual-arm cases. The physical experiment further demonstrates the planning-execution workflow of \drafto for complex tasks, such as object grabbing from a drawer.

\section{Related Works}
\label{sec:related}

\subsection{Waypoint-based trajectory optimization}
Most optimization-based planners work directly on discrete waypoints. CHOMP \cite{zucker_chomp_2013} updates the trajectory with functional gradients, TrajOpt \cite{schulman_motion_2014} solves a sequence of convexified subproblems, and GPMP2 \cite{mukadam_continuous-time_2018} uses a Gaussian-process prior with dense interpolation between support states. These methods are effective, but their continuous-time safety and joint-limit handling still become more expensive as the waypoint density increases. Similar issues also arise in SQP- and MPC-style formulations for constrained motion and coordination \cite{woolfrey_predictive_2021, kabir_generation_2021}, where trajectory-wide inequalities necessitate repeated linearization and constraint checking. Warm-start strategies \cite{lembono_memory_2020} and multimodal search \cite{osa_multimodal_2020} can improve robustness, but they do not eliminate the burden of dense discretization or the need for repeated constrained subproblems.

\subsection{Function-space trajectory parameterization}
Another direction is to optimize a continuous trajectory through a low-dimensional parameterization. GP-based models keep a continuous-time view of the motion \cite{mukadam_continuous-time_2018, huang_motion_2017}, and recent extensions improve inference flexibility or differentiability \cite{cosier_unifying_2024, bhardwaj_differentiable_2020, li_constrained_2025}. Basis-function parameterization \cite{feng_optimal_2023} uses a small set of coefficients to control the whole trajectory, so local coefficient updates produce global motion changes. FACTO \cite{feng_facto_2026} follows this idea by optimizing the trajectory-wide constrained objective in coefficient space. \drafto adopts the same function-space representation but addresses a different issue: how to keep the main iterations efficient when the number of constraints remains large.

\subsection{Globalization and feasibility handling}
From the solver side, constrained motion planning has relied on Newton and quasi-Newton updates \cite{nocedal_numerical_2006}, trust-region methods \cite{schulman_trust_2015}, and constrained QP backbones \cite{badreddine_sequential_2014}, often combined with LM-style regularization \cite{levenberg_method_1944} or operator-splitting solvers such as OSQP and ADMM \cite{boyd_distributed_2011, stellato_osqp_2020}. 
Stochastic trajectory optimizers, such as STOMP \cite{kalakrishnan_stomp_2011}, SMTO \cite{osa_multimodal_2020}, and STO-GPs \cite {petrovic_stochastic_2019}, improve exploration in nonconvex problems, but typically require many rollouts. 
Also, SGD methods \cite{bottou_-line_1999, kingma_adam_2014} are integrated to solve large-scale trajectory optimization problems \cite{feng_incrementally_2022}. 
In practice, when strict inequality constraints are enforced at every iteration, the constrained subproblem can become the main computational bottleneck. The method proposed here addresses that bottleneck: instead of solving a constrained QP throughout the full optimization, \drafto uses reduced-space Gauss--Newton steps for the main search and reserves the constrained QP only for initialization and terminal feasibility repair.

\section{Preliminary}
\subsection{Function-space trajectory representation}
\label{sec:facto}

Considering the trajectory-wide safety and task feasibility, this work still utilizes the function-space optimization framework proposed by FACTO \cite{feng_facto_2026}. In the same way, a scalar trajectory $\xi$ with one DoF over time interval $[0, T]$ can be parameterized by a truncated expansion of $N$ orthogonal basis functions $\bm\phi(t)$ combined with a boundary lift $\varphi(t)$:
\begin{equation}
\begin{gathered}
    \xi(t) = \bm\phi(t)^\top \bm\psi +  \varphi(t), \\[2pt]
    \quad \bm\phi(t) := [\phi_0(t),\dots,\phi_N(t)]^\top \in \R^{N+1},
    \quad \bm\psi \in \R^{N+1}.
\end{gathered}
\label{eq:fs_scalar}
\end{equation}
It can be extended to the $M$-DoF trajectory $\bm\xi$ by stacking one-DoF basis functions: 
\begin{equation} \label{eq:fs_vector}
\begin{gathered}
    \bm\xi(t) = \bm\Phi(t)\bm\psi + \bm{\varphi}(t), \quad 
    \dot{\bm\xi}(t) = \dot{\bm\Phi}(t)\bm\psi + \dot{\bm{\varphi}}(t). 
\end{gathered}
\end{equation}
where the basis stack $\bm\Phi(t)\ =\ \mathbf I_{M \times M} \otimes \bm\phi^\top \in \mathbb{R}^{M\times M(N+1)}$ and the extended coefficient $\bm\psi \in \R^{M(N+1)}$.  

\subsection{Problem Setup and Costs}
\label{sec:problem}

The objective form of GPMP2 \cite{mukadam_continuous-time_2018} with a parameter $\varrho$  is adopted to balance between trajectory smoothness and safety: 
\begin{equation} \label{eq:obj}
    \mathcal{F}(\bm \psi) 
    = \varrho\,\mathcal{F}_{\mathrm{smooth}}(\bm \psi)\;+\;\mathcal{F}_{\mathrm{obs}}(\bm \psi). 
\end{equation}
where the smoothness term is calculated by 
\begin{equation} \label{eq:smooth}
    \mathcal{F}_{\mathrm{smooth}}(\bm \psi)=\bm \psi^\top \mathbf Q\,\bm \psi, \,\,\,\,
    \mathbf Q = \| \dot{\bm\Phi}(t) \|_{w}^{2},
\end{equation}
with a positive weight function $w(t)$ for the Lebesgue integral, and the obstacle term is calculated by 
\begin{equation} \label{eq:obs0}
    \begin{aligned}
        \mathcal{F}_{\mathrm{obs}}(\bm \psi)
        = \int_{0}^{T} f_o(\bm\psi) \diff t 
        = \int_{0}^{T} \sum_{\mathcal B_i \in \bm{\mathcal B}} |d(\mathcal B_{i,t})|^+ \diff t,  
    \end{aligned}
\end{equation}
with the $\ell_1$-penalty of the signed distance $d$ of all the collision-check balls (CCBs, $\bm{\mathcal B}$).  
The self- and cross-collision checks of \cite{feng_facto_2026} are used for multi-robot trajectory optimization. 

Since the obstacle term is non-convex and analytically non-integrable, a discrete squared-integral form is adopted to approximate the objective of \eqref{eq:obj}: 
\begin{equation} \label{eq:work_obj}
    \begin{aligned}
        \widehat{\mathcal F}(\bm \psi)
        \;=\; & \varrho \,\bm \psi^\top\mathbf Q\,\bm \psi \;+\; \sum_k f_o\!\big(\bm\Phi(t_k)\bm \psi\big)^2 \\
        \;\xlongequal{\Delta\bm\psi \rightarrow 0}\;&  \varrho\,\|\bm \psi + \Delta\bm\psi\|_{\mathbf Q}^2 \;+\; \sum_{k} \big( r_k(\bm\psi) + \mathbf{g}_k^\top \Delta \bm\psi \big)^2, 
    \end{aligned} 
\end{equation}
where $\Delta\bm\psi$ is the update step of $\bm\psi$, and $r_k(\bm\psi) = f_o\!\big(\bm\Phi(t_k)\bm \psi\big)$ and $\mathbf{g}_k$ denote the collision-cost residual and its gradient, respectively. 
Then, we flatten the gradient $\widehat{\mathbf g} = \sum r_k \mathbf{g}_k$  and Hessian $\widehat{\mathbf{H}} = \sum \mathbf{g}_k \mathbf{g}_k^\top$ of  the squared-integral obstacle function by unbiased exponential moving averaging (uEMA) and get the simplified objective model with the flattened gradient $\overline{\mathbf{g}}$  and Hessian $\overline{\mathbf{H}}$: 
\begin{equation} \label{eq:model}
    \begin{aligned}
        m(\Delta\bm \psi)
        = \Delta\bm \psi^\top\! (\varrho \mathbf Q+ \overline{\mathbf H}) \Delta\bm \psi 
        + 2(\varrho \mathbf{Q} \bm\psi+ \overline{\mathbf g})^\top \Delta\bm \psi. 
    \end{aligned}
\end{equation}

\subsection{Constraint Construction}
\label{sec:constr}

This section presents a method for constructing the boundary, task, and joint-limit constraints. 

\subsubsection{Boundary constraints}
Given the initial and goal waypoints $\bm\theta_0, \bm\theta_g$,  the boundary constraints are built as
\begin{equation} \label{eq:boundary_constr}
\begin{gathered}
    \bm\Phi(0)\,\bm \psi=\bm\theta_0 - \bm\varphi(0),\quad
    \bm\Phi(T)\,\bm \psi=\bm\theta_g - \bm\varphi(T). 
\end{gathered}
\end{equation}

\subsubsection{Task constraints} 
Given a task requirement $\bm\chi  = [\bm x_{\min}, \bm x_{\max}]$ and current posture $\bm x$ , the desired posture can be calculated by $\bm x_\text{des} = \min(\max(\bm x, \bm x_{\min}), \bm x_{\max})$. Both $\bm x$ and $\bm x_\text{des}$ can be further expressed as $\bm\zeta$  and $\bm\zeta_\text{des}$ in  $\mathfrak{se}(3)$, respectively. And the trajectory-wide task constraint is 
\begin{equation} \label{eq:task_constr_traj}
    \bm{h}(t) := \big(\bm{\zeta}(\bm\xi) - \bm{\zeta}_{\mathrm{des}}(\bm\xi) \big)(t) = \mathbf 0, \,\,
    \bm\xi(t) = (\bm \varphi + \bm\Phi\bm\psi)(t). 
\end{equation}
Then, the discrete trajectory-wide constraint becomes
\begin{equation} \label{eq:task_constr}
    \begin{gathered}
        \bm{\mathcal H}(\bm \psi) : = \big[\,\bm h_{1}^\top \;\dots \;\bm h_{K_\text{tsk}}^\top \, \big]^\top=\bm 0, \;
    \bm h_{k} = \bm h\big(\bm\Phi(t_{k})\bm \psi \big). 
    \end{gathered}
\end{equation}

\subsubsection{Joint limit constraints}
Given the joint-position limits $[\bm\theta_{\min}, \bm\theta_{\max}]$, the discrete inequality constraints are  
\begin{equation} \label{eq:joint_limit}
    \bm\theta_{\min}^{\oplus K_{\text{lmt}}} 
    \leq \mathbf{A}_\mathrm{ieq} \bm\psi 
    \leq \bm\theta_{\max}^{\oplus K_{\text{lmt}}}, 
\end{equation}
where $K_{\text{lmt}}$ is the number of limit-checking points and $\mathbf{A}_\mathrm{ieq} = [\bm\Phi(t_1)^\top \dots \bm\Phi(t_{K_\text{lmt}})^\top]^\top$. Then, the inequality constraint of each update step $\Delta\bm\psi$ becomes
\begin{equation}\label{eq:ieq_constr_stack}
    \mathbf{b}_\mathrm{lo} \leq \mathbf{A}_\mathrm{ieq} \Delta\bm\psi \leq \mathbf{b}_\mathrm{up}, 
\end{equation}
where $\mathbf{b}_\mathrm{lo} = \bm\theta_{\min}^{\oplus K_{\text{lmt}}} - \mathbf{A}_\mathrm{ieq} \bm\psi$ and $\mathbf{b}_\mathrm{up} = \bm\theta_{\max}^{\oplus K_{\text{lmt}}} - \mathbf{A}_\mathrm{ieq} \bm\psi$.

\subsection{Reduced-space Formulation}
\label{sec:reduced}
Boundary \eqref{eq:boundary_constr} and task \eqref{eq:task_constr_traj} constraints are the only two types of equality constraints in our trajectory optimization. This work uses $\mathbf A_{\mathrm{eq}}$ and $\mathbf{b}_{\mathrm{eq}}$ to stack their linearized form: 
\begin{equation} \label{eq:eq_constr_stack}
    \begin{gathered}
        \mathbf A_{\mathrm{eq}}\,\Delta\bm \psi\;=\;\mathbf{b}_{\mathrm{eq}},\\[6pt]
        \mathbf A_{\mathrm{eq}} =
        \begin{bmatrix}
            [\bm\Phi^\top(0), \, \bm\Phi^\top(T)]^\top \\
            \partial_{\bm\psi} \bm{\mathcal H}
        \end{bmatrix},
        \,
        \mathbf{b}_{\mathrm{eq}} =
        -\begin{bmatrix}
            \bm 0 \\
            \bm{\mathcal H}
        \end{bmatrix}.
    \end{gathered}
\end{equation} 
Then, the step updating in the null space $\mathbf{N}$ of $\mathbf A_{\mathrm{eq}}$ becomes 
\begin{equation} \label{eq:null_update}
    \bm\psi \leftarrow \bm\psi + \Delta\bm\psi_0 + \mathbf{N}\bm z
\end{equation} 
with a feasible initial step $\Delta\bm\psi_0$ and a null-space step $\bm z$. 

\section{\drafto: Decoupled Reduced-Space Search and Feasibility Repair Trajectory Optimization}
\label{sec:method}

This section introduces how \drafto (\algref{alg:drafto}) resolves the low efficiency, mainly caused by a large amount of inequality constraints for the trajectory-wide joint limits. 
It separates \textit{feasibility enforcement} for the joint-limit and \textit{step searching} for the constrained collision-free trajectory optimization. The \textit{feasibility enforcement} is only performed on the initial and terminal stages to strictly meet the boundary, task, and joint-limit constraints. The \textit{step searching} is performed during the main iterations by computing the reduced-space GN steps with adaptive quasi-Hessian regularization. 

\begin{algorithm}[!htp]
\caption{\drafto}
\label{alg:drafto}
\DontPrintSemicolon
\KwIn{Initial coefficients $\bm\psi_0$, basis $\bm\Phi(t)$, null-space basis $\mathbf N$, checkpoint count $K_{\text{chk}}$, tolerances}
\KwOut{Optimized coefficients $\bm\psi^\star$}
Initialize $\bm\psi \leftarrow \bm\psi_0$ via \eqref{eq:qp_init}, regulator $\lambda \leftarrow \lambda_0$\;
\For{$i=1$ \KwTo $N_{\max}$}{
    Rebuild sampled obstacle residuals, task linearization, and the local model \eqref{eq:model}; \;
    Build soft joint-limit penalty at checkpoints \eqref{eq:drafto_soft_jl}; \;
    Form the reduced gradient/Hessian and solve the damped system \eqref{eq:lm_red}; \;
    Recover $\Delta\bm\psi = \Delta\bm\psi_0 + \mathbf N \bm z$ and run line search; \;
    Accept the step $\Delta\bm\psi$ using the two-phase rule; \;
    Update $\bm\psi$, regulator $\lambda$, and stopping statistics; \;
    \lIf{converged}{break; }
}
Evaluate terminal checkpoint violation $v_\infty(\bm\psi)$\;
\If{$v_\infty(\bm\psi)>\varepsilon_{\mathrm{jnt}}$}{
    Rebuild $\mathbf A_{\mathrm{eq}},\mathbf b_{\mathrm{eq}},\mathbf A_{\mathrm{ieq}},\mathbf b_\mathrm{lo},\mathbf b_\mathrm{up}$; \;
    Solve constrained QP \eqref{eq:qp_terminal} for feasibility repair; \;
}
\Return{$\bm\psi^\star \leftarrow \bm\psi$}. \;
\end{algorithm}

\subsection{Initialization and equality-feasible starting point}
\label{sec:init_qp}

\drafto performs the constrained QP for the initialization of the trajectory optimizer by only considering the trajectory smoothness \eqref{eq:smooth} and boundary conditions \eqref{eq:boundary_constr}: 
\begin{equation} \label{eq:qp_init}
    \begin{aligned}
        \bm\psi_0 &= \argmin_{\bm\psi}\; \bm\psi^\top \mathbf{Q} \bm\psi \\
        & \text{s.t.}\; \text{boundary constraints in \eqref{eq:boundary_constr}.}
    \end{aligned}
\end{equation}
Since the matrix $\mathbf{Q}$, due to the orthogonality, is strictly diagonal, the whole constrained-QP problem is highly sparse and can be solved fast. This initialization process can not only ensure a stable warm start but also accelerate subsequent optimization, especially for unconstrained tasks, because the null space of the boundary constraints is constant.

\subsection{Soft joint-limit penalty in coefficient space}
\label{sec:soft_jl}
Although the null-space method can reduce the optimization scale and remove equality constraints, the number of inequality constraints remains large, thereby consuming substantial computational resources. Therefore, the joint-limit violations are penalized, and the pure GN instead of the constrained QP is employed to elevate the computation efficiency by eliminating the inequality constraints. 

In this way, we first transfer the double-sided inequality stack of \eqref{eq:ieq_constr_stack} into a single-sided stack: 
\begin{equation}
    \widetilde{\mathbf A}
    :=
    \begin{bmatrix}
        \mathbf A_\mathrm{ieq} \
        -\mathbf A_\mathrm{ieq}
    \end{bmatrix},
    \qquad
    \widetilde{\bm b}
    :=
    \begin{bmatrix}
        \mathbf b_\mathrm{up} \
        -\mathbf b_\mathrm{lo}
    \end{bmatrix}.
    \label{eq:one_sided_stack}
\end{equation}
And the joint-limit violation vector is defined as
\begin{equation}
    \bm v(\bm\psi)
    :=  \widetilde{\mathbf A}\bm\psi - \widetilde{\bm b}
    \;\xlongequal{\Delta\bm\psi \rightarrow 0} \widetilde{\mathbf A}(\bm\psi + \Delta\bm\psi) - \widetilde{\bm b}.
    \label{eq:violation_vec}
\end{equation}
Then, a hinge-squared penalty, activated by the violated rows or inequalities, can be expressed as
\begin{equation}
    \widehat{\mathcal F} _{\mathrm{jnt}}(\bm{\psi})
    :=
    \frac{1}{\sigma_{\mathrm{jnt}}^2}
    \sum_{k=1}^{K_\mathrm{jnt}}
    \big[\max(0, v_k(\bm{\psi}))\big]^2,
    \label{eq:drafto_soft_jl}
\end{equation}
where $\sigma_{\mathrm{jnt}}>0$ controls the penalty strength and $K_\mathrm{jnt}$ is the number of joint-limit checkpoints. 

To improve the computation efficiency of penalty update, especially when the checkpoints are dense, the active index set $\mathcal I_{\mathrm{jnt}}:=\{k\mid v_k(\bm\psi)>0\}$ is formed such that only a small number of violated checkpoints are considered at each QP step.  In this way, the corresponding GN terms are
\begin{equation}
    \mathbf g_{\mathrm{jnt}}
    =
    \sigma_{\mathrm{jnt}}^{-2}{\sum_{k\in\mathcal I_{\mathrm{jnt}}} v_k\,\tilde{\bm a}_k},
    \quad
    \mathbf H_{\mathrm{jnt}}
    \approx
    \sigma_{\mathrm{jnt}}^{-2}\sum_{k\in\mathcal I_{\mathrm{jnt}}} \tilde{\bm a}_k\tilde{\bm a}_k^\top, 
    \label{eq:soft_jl_gn_terms}
\end{equation}
where $\tilde{\bm a}_i^\top$ is the $i$-th row of $\widetilde{\mathbf A}$. 

The next section introduces how the damped GN search works based on a modified working objective combining the soft joint-limit penalty \eqref{eq:drafto_soft_jl}: 
\begin{equation} \label{eq:work_obj_mod}
    \mathcal J(\bm\psi) = \widehat{\mathcal F}(\bm\psi) + \widehat{\mathcal F}_\mathrm{jnt}(\bm\psi). 
\end{equation}

\subsection{Damped Gauss--Newton search in the reduced space}
\label{sec:pure_gn}
Based on the modified working objective \eqref{eq:work_obj_mod}, combining the simplified objective model \eqref{eq:model} and soft joint-limit terms \eqref{eq:soft_jl_gn_terms} yields a full-space gradient and Hessian approximation:
\begin{equation}
    \mathbf g := \varrho \mathbf Q\bm\psi + \overline{\mathbf g} + \mathbf g_{\mathrm{jnt}},
    \quad
    \mathbf H := \varrho \mathbf Q + \overline{\mathbf H} + \mathbf H_{\mathrm{jnt}}.
    \label{eq:full_model_terms}
\end{equation}
The equality-feasible parameterization \eqref{eq:null_update} converts the step to the reduced coordinate $\bm z$. Substituting it into \eqref{eq:model} gives
\begin{equation}
    m_{\mathrm{red}}(\bm z)
    :=
    \bm z^\top \mathbf H_{\mathrm{red}} \bm z + 2\,\mathbf g_{\mathrm{red}}^\top \bm z,
    \label{eq:red_model}
\end{equation}
with
\(
    \mathbf H_{\mathrm{red}} = \mathbf N^\top \mathbf H \mathbf N
\) 
and 
\(
    \mathbf g_{\mathrm{red}} = \mathbf N^\top\!\big(\mathbf H\Delta\bm\psi_0 + \mathbf g\big).
    \label{eq:red_model_terms}
\)
Then, the above GN model in a reduced/null space can be solved with a regularized Hessian matrix:
\begin{equation}
    \big(\mathbf H_{\mathrm{red}}+\lambda \mathbf I\big)\bm z = -\mathbf g_{\mathrm{red}},
    \label{eq:lm_red}
\end{equation}
where $\lambda>0$ is adapted across iterations based on the trust-region method \cite{byrd_trust_2000, nocedal_numerical_2006}.  We first define the model and actual reductions as 
\[
\textit{mred} 
= \mathbf g_{\mathrm{red}}^\top [2 \mathbf{I} - \big(\mathbf H_{\mathrm{red}}+\lambda \mathbf I\big)^{-1} \mathbf H_{\mathrm{red}}]\, \big(\mathbf H_{\mathrm{red}}+\lambda \mathbf I\big)^{-1} \mathbf g_{\mathrm{red}}
\]
\[
\textit{ared} = \mathcal J(\bm\psi) - \mathcal J(\bm\psi + \Delta\psi_0 + \mathbf{N}\bm z), 
\]
Then the reduction fraction $\textit{ared}/\textit{mred}$ is used to adjust the $\lambda$ adaptively: if the fraction is beyond the upper threshold, $\lambda$ will be decreased to recover a more aggressive GN step; if the reduction fraction is below the lower threshold, meaning the reduction system is poorly conditioned, $\lambda$ will be increased. However, the GN step can still perform poorly even when $\lambda$ reaches its maximum value $\lambda_{\max}$. Therefore, the next section introduces the globalization rule for stable convergence.

\subsection{Globalization with a two-phase acceptance rule}
\label{sec:nonmono}

The modified objective \eqref{eq:work_obj_mod} shows that the obstacle term changes sharply when the active CCB switches and that the soft joint-limit term is disturbed severely when the violated inequality changes with a small $\sigma_\mathrm{jnt}$. This means that there always exist temporary non-descent steps from an infeasible solution to a feasible one. Therefore, \drafto uses a two-phase acceptance rule for the full coefficient update:  
\begin{equation}
    \Delta\bm\psi = \Delta\bm\psi_0 + \mathbf N\bm z,
    \qquad
    \bm\psi \leftarrow \bm\psi + \alpha \Delta\bm\psi,
    \label{eq:coeff_update}
\end{equation}
with step length $\alpha\in(0,1]$.

\paragraph*{Phase I (exploratory stage)}
In the early iterations, the GN step is executed aggressively with LM damping and $\alpha=1$. 
The reduction ratio $\textit{ared}/\textit{mred}$ is only used to update $\lambda\in[\lambda_{\min},\lambda_{\max}]$, not for step acceptance. 

\paragraph*{Phase II (tail stabilization)}
Once the trajectory approaches feasibility, a non-monotone acceptance rule is used to prevent stagnation while ensuring descent relative to the most recent iterations.
Define the windowed reference
\begin{equation}
\bar{\mathcal J}_i := \max\{\mathcal J_{i-j}\mid 0\le j\le W-1\}.
\label{eq:nonmono_ref}
\end{equation}
For a candidate step $\Delta\bm\psi$ and step size $\alpha$, we accept if
\begin{equation}
    \mathcal J(\bm\psi_i+\alpha\Delta\bm\psi)
    \le
    \bar{\mathcal J}_i - c_1 \alpha\, \Delta\bm\psi^\top \mathbf g,
    \label{eq:nonmono_accept}
\end{equation}
with $c_1\in(0,1)$. 
If the test fails, we backtrack by setting $\alpha\leftarrow \beta\alpha$ ($\beta\in(0,1)$) until acceptance or the maximum number of trials is reached. 
This tail rule allows small temporary increases while requiring descent relative to the recent window.

\subsection{Terminal-stage hard-feasibility repair}
\label{sec:final_fix}

Although the above soft penalty method drives the entire trajectory optimization away from joint limits, it cannot ensure strict joint-limit feasibility at termination. Therefore, once the termination condition is met, \drafto first performs a strict feasibility check by 
\begin{equation}
    \begin{aligned}
        v_\infty(\bm{\psi})
        :=
        \max\!\Big(
        & \|\max(\bm 0,\; \mathbf A_\mathrm{ieq}\bm\psi - \bm\theta_{\max}^{\oplus K_{\text{lmt}}})\|_\infty, \\
        & \|\max(\bm 0,\; \bm\theta_{\min}^{\oplus K_{\text{lmt}}} - \mathbf A_\mathrm{ieq}\bm\psi)\|_\infty
        \Big). 
    \end{aligned}
    \label{eq:vinf}
\end{equation}
If $v_\infty(\bm\psi)\le \varepsilon_{\mathrm{jnt}}$, the result is accepted. Otherwise, \algref{alg:drafto} uses the same local model \eqref{eq:model} and the current linearized equalities \eqref{eq:eq_constr_stack} to solve a constrained QP: 
\begin{equation} \label{eq:qp_terminal}
    \begin{aligned}
        \Delta\bm\psi^\star = 
        & \argmin_{\Delta\bm\psi}\;
        m(\Delta\bm\psi) + \lambda_\mathrm{reg} \|\Delta\bm\psi\|^2 \\
        \text{s.t.} \quad
        & \mathbf A_{\mathrm{eq}}\Delta\bm\psi=\mathbf b_{\mathrm{eq}}, \;
        \mathbf b_\mathrm{lo}\le \mathbf A_\mathrm{ieq}\Delta\bm\psi \le \mathbf b_\mathrm{up}, 
    \end{aligned}
\end{equation}
where a large regulator $\lambda_\mathrm{reg}$ ensures the final repaired solution is still near the local optimum. 
Although the constrained QP step with a dense joint-limit check consumes more computational resources than the pure GN step, its overall computational effort is still negligible because only one or a small number of repair iterations are required in practice.

\subsection{Dense and configurable checkpointing}
\label{sec:checkpoint}
The finite set of discrete waypoints is used to verify trajectory-wide constraints, given the non-existence of an analytic integral solution. Therefore, \drafto utilizes the dense uniform checkpoints for the joint-limit constraints: 
\begin{equation}
    t_k = \frac{k-1}{K_{\mathrm{chk}}-1}T, \qquad k=1,\dots,K_{\mathrm{lmt}}, 
    \label{eq:checkpoints}
\end{equation}
where a large $K_\mathrm{lmt}$ is selected for the dense checking during the {feasibility repair} and a smaller $K_\mathrm{check}$ is selected for the sparser checking during the two-phase GN process. Additionally, \drafto selects a few significant task-constraint checkpoints to rebuild the task constraints of \eqref{eq:task_constr}:
\begin{equation}
    \begin{gathered}
        \bm{\mathcal H}(\bm \psi) : = \big[\,\bm h_{\pi(1)}^\top \;\dots \;\bm h_{\pi(K_\text{tsk})}^\top \, \big]^\top=\bm 0, \\[2pt]
        \text{with }\{ \pi |  \|h_{\pi(i)}\| \geq  \|h_{\pi(i+1)}\|, \, i =1\dots N_\mathrm{tsk}  \}. 
    \end{gathered}
\end{equation}

\subsection{Convergence Discussion}

\begin{thm}[Local Convergence]
\label{thm:drafto_conv}
According to the definition of the modified working objective $\mathcal{J}$ in \eqref{eq:work_obj_mod}, let $\mathcal L := \{\bm\psi \mid \mathcal J(\bm\psi)\le \mathcal J(\bm\psi_0)\}$ be the level set and assume: (i) $\mathcal J$ is bounded below and $\nabla \mathcal J$ is Lipschitz continuous; (ii) $\mathbf A_{\mathrm{eq}}$ has constant rank (ensuring a well-defined null-space $\mathbf N$); (iii) $\mathbf H_{\mathrm{red},k}+\lambda_k \mathbf I \succeq \mu \mathbf I$ for some $\mu > 0$; (iv) Line search steps satisfy the acceptance rule \eqref{eq:nonmono_accept}. 

Then, the iterates satisfy: (i) Every iterate remains feasible for the linearized equality system; (ii) $\{J(\bm\psi_k)\}$ is bounded and convergent; (iii) Any accumulation point $\bar{\bm\psi}$ satisfies $\mathbf N(\bar{\bm\psi})^\top \nabla J(\bar{\bm\psi}) = \bm 0$. 
If LICQ holds and the terminal problem \eqref{eq:qp_terminal} is locally feasible, the repair stage reaches checkpoint feasibility in finitely many steps.
\end{thm}

\section{Experiment and Result Analysis}

\subsection{Benchmarks and setup}
\label{sec:setup}

\drafto is implemented on FR3 to benchmark against 4 optimization-based planners, such as  FACTO \cite{feng_facto_2026}, CHOMP \cite{zucker_chomp_2013}, TrajOpt \cite{schulman_finding_2013}, and GPMP2 \cite{mukadam_continuous-time_2018}, as well as 3 sampling-based planners, such as RRT-C \cite{kuffner_rrt-connect_2000}, RRT* \cite{karaman_sampling-based_2011}, and PRM \cite{kavraki_probabilistic_1996}. 
All the benchmark tests are performed across 1000 planning tasks in 10 single-arm planning scenes and 80 tasks in 2 dual-arm scenes. 
In addition, we replicate all optimization-based planners and replace OMPL's feasible-validation module with Pinocchio-FCL to ensure that all planners use the same kinematics library, Pinocchio. All tests are performed on a single-threaded 2.3 GHz Intel Core i9 with 4GB of RAM. 

\subsubsection{Parameter settings}
For the optimization-based planners, we use the same safety margin for collision checks, the same termination condition, and the same smoothness weight. Both FACTO and \drafto share the setting of the basis number $N$, basis function, and the collision checkpoint number. $N$ is also used by GPMP2 for the number of support waypoints. The doubled checkpoint numbers are used by all optimization-based planners except FACTO and \drafto to ensure time-continuous, collision-free operation. For the sampling-based planners, we use the default settings and a 10-second time budget. To test the constrained tasks, RRT-C adopts the IMACS \cite{kingston_exploring_2019} framework by applying a projection-based constrained space, whereas CHOMP uses the Lagrangian multiplier method.

\subsubsection{Baselines and ablations}
\label{sec:baselines}

Since \drafto decouples the reduced-space GN search from the constrained QP and employs a two-stage globalization policy, we use FACTO, which uses the constrained QP throughout the optimization, as the baseline to demonstrate the advantage of the decoupled process. We also use \drafto-GN, which performs the pure GN steps without a non-monotone tail policy, as an ablation study to assess the effects of the two-stage policy.

\subsubsection{Metrics}
\label{sec:metrics}

The success rate (S in \%), computation time (T in seconds), and trajectory roughness (R in rad/s$^2$) are used as metrics for the benchmark test.  A dense collision check is used to determine the trajectory’s success. Since the end-effector fingers attach to the obstacle in some planning tasks, the collision checks for the fingers are deprecated in this work. The computation time includes the entire optimization process from initialization to feasibility repair. The normalized squared joint-acceleration accumulation is used for roughness: 
\begin{align*}
    \emph{roughness}  
    &= \frac{1}{K-1} \sum_{k=1}^{K-1}  \frac{\|\bm\theta_{k-1} - 2 \bm\theta_{k} + \bm\theta_{k+1}\|}{(t_k - t_{k-1})^2} 
\end{align*}
where $\bm\theta_k = \bm\Psi(t_k) \bm\psi$ and $t_k = {k}/{(K-1)}$. 

In addition, we report only their means and maximum values (avg/max) because the T \& R distributions are non-normal.

\subsection{Benchmark results and analysis}
\label{sec:results}

\subsubsection{Single-arm case}

According to the density of the obstacle surrounding the initial trajectory, the distance between the trajectory and joint limits, and the number of constraints, we sort the single-arm planning difficulty as: \emph{kitchen\_constr} > \emph{cage} = \emph{table\_under\_pick\_constr} > \emph{table\_under\_pick} = \emph{kitchen} > \emph{box} > \emph{table\_pick} > \emph{bookshelf}. Therefore, we focus on analyzing the results of \emph{cage}, \emph{kitchen} and \emph{table\_under\_pick}, shown in \tabref{tab:single_arm}, and the results of \emph{kitchen\_constr} and \emph{table\_under\_pick\_constr}, shown in \tabref{tab:single_arm_constr}. The results of the others are placed in \tabref{tab:benchmark_appendix_scenes}. 

\begin{figure}[!htp]
    \centering
    \hfill
    \begin{subfigure}[t]{1\linewidth}
        \centering
        \includegraphics[width=0.32\linewidth]{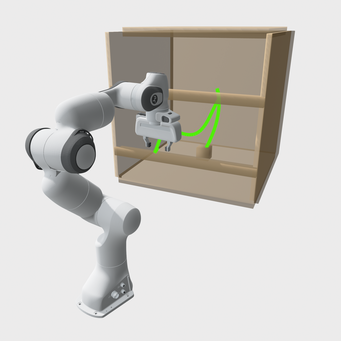}
        \includegraphics[width=0.32\linewidth]{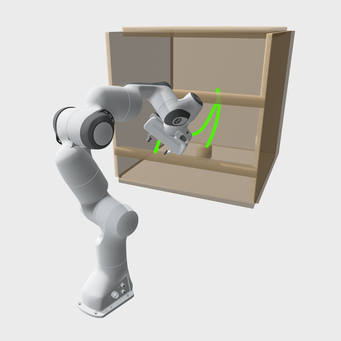}
        \includegraphics[width=0.32\linewidth]{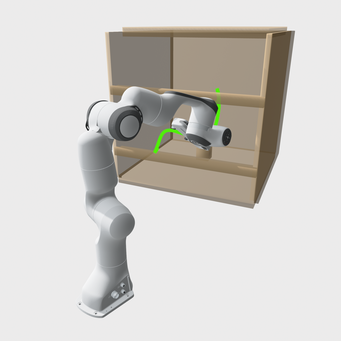}
        \\[2pt]
        \includegraphics[width=0.32\linewidth]{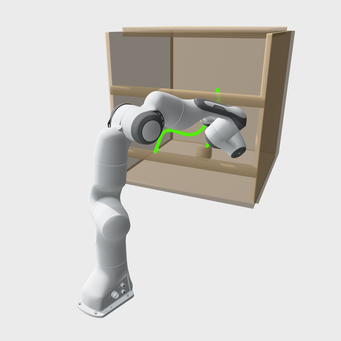}
        \includegraphics[width=0.32\linewidth]{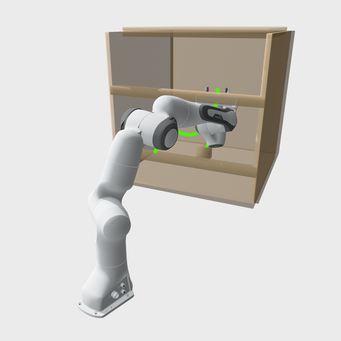}
        \includegraphics[width=0.32\linewidth]{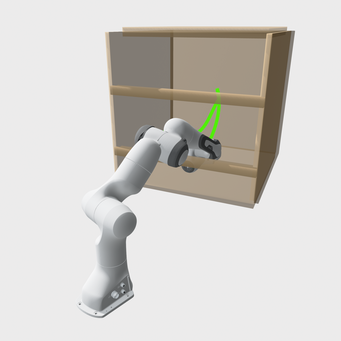}
    \end{subfigure}
    \hfill
    \caption{Simulation result of the single FR3 arm in \emph{cage}: representative snapshots along the trajectory are shown (green). }
    \label{fig:simu_single_arm_01}
\end{figure}

\begin{table*}[!htp]
\centering
\setlength{\tabcolsep}{2.5pt}
\caption{Benchmark results across single-arm planning scenes.}
\label{tab:single_arm}
\begin{tabular}{@{}llccccccccc@{}}
\toprule
Scene & Metrics & \drafto-GN & \drafto & FACTO & CHOMP & TrajOpt & GPMP2 & RRT-C & RRT* & PRM \\
\midrule

\multirow{3}{*}{cage}
& S (\%)        & 96 & 96 & 95 & 19 & 0  & 82  & 92   & 0  & 17 \\
& T (avg/max)   & 0.031\,/\,0.167 & 0.030\,/\,0.218 & 0.134\,/\,1.338 & 0.155\,/\,0.235 & - & 0.213\,/\,1.545 & 3.779\,/\,10.006 & - & 4.538\,/\,9.904 \\
& R (avg/max)   & 86.9\,/\,112.0  & 87.2\,/\,118.3  & 91.1\,/\,154.5  & 47.9\,/\,55.3   & - & 48.2\,/\,107.4  & 386.3\,/\,1322.5 & - & 139.5\,/\,435.9 \\
\midrule

\multirow{3}{*}{kitchen}
& S (\%)        & 92 & 92 & 94 & 85 & 43 & 74 & 99 & 55 & 78 \\
& T (avg/max)   & 0.028\,/\,0.269 & 0.028\,/\,0.147 & 0.084\,/\,0.929 & 0.159\,/\,0.438 & 0.225\,/\,0.526 & 0.113\,/\,0.200 & 0.538\,/\,6.024 & 1.638\,/\,8.796 & 1.233\,/\,8.206 \\
& R (avg/max)   & 79.5\,/\,135.3  & 79.5\,/\,135.3  & 79.3\,/\,154.4  & 53.7\,/\,115.5  & 299.3\,/\,212.5 & 33.0\,/\,46.4   & 189.5\,/\,824.7 & 149.7\,/\,506.7 & 274.4\,/\,630.5 \\
\midrule

\multirow{3}{*}{\parbox{1.25cm}{table\\\_under\\\_pick}}
& S (\%)        & 97 & 97 & 98 & 36 & 56 & 83 & 98 & 16 & 84 \\
& T (avg/max)   & 0.032\,/\,0.102 & 0.030\,/\,0.075 & 0.060\,/\,0.207 & 0.159\,/\,0.237 & 0.208\,/\,0.436 & 0.091\,/\,0.184 & 0.238\,/\,5.670 & 2.204\,/\,9.810 & 1.855\,/\,10.003 \\
& R (avg/max)   & 108.2\,/\,414.8 & 108.2\,/\,414.8 & 85.4\,/\,137.1  & 45.8\,/\,63.5   & 340.5\,/\,771.2 & 34.3\,/\,53.8   & 144.7\,/\,377.2 & 138.9\,/\,298.9 & 134.1\,/\,553.5 \\
\bottomrule
\end{tabular}
\end{table*}

\begin{table*}[!htp]
\centering
\caption{Constrained-task benchmark results across single-arm planning scenes.}
\label{tab:single_arm_constr}
\begin{tabular}{@{}llccccc@{}}
\toprule
Scene & Metrics & \drafto-GN & \drafto & FACTO & CHOMP & RRT-C \\
\midrule

\multirow{3}{*}{kitchen\_constr}
& S (\%)      & 89 & 88 & 90 & 10 & 86 \\
& T (avg/max) & 0.073\,/\,0.243 & 0.068\,/\,0.247 & 0.208\,/\,1.027 & 1.592\,/\,1.807 & 4.036\,/\,8.782 \\
& R (avg/max) & 126.6\,/\,286.1 & 124.4\,/\,286.1 & 141.2\,/\,424.3 & 335.5\,/\,1852.2 & 2117.2\,/\,7570.2 \\
\midrule

\multirow{3}{*}{\parbox{2cm}{table\_under\\\_pick\_constr}}
& S (\%)      & 95 & 95 & 96 & 9 & 72 \\
& T (avg/max) & 0.053\,/\,0.290 & 0.047\,/\,0.114 & 0.177\,/\,0.984 & 1.492\,/\,1.834 & 4.185\,/\,10.056 \\
& R (avg/max) & 128.9\,/\,314.8 & 128.8\,/\,314.8 & 177.6\,/\,472.1 & 347.1\,/\,2043.6 & 2297.1\,/\,5229.9 \\
\bottomrule
\end{tabular}
\end{table*}

\paragraph{Comparison across the algorithms}

\tabref{tab:single_arm} and \tabref{tab:single_arm_constr} both show that \drafto takes 40\%-75\% less computation time than FACTO and just has a tiny success-rate drop. It suggests that decoupling the constrained QP and penalized GN descent processes can significantly improve efficiency while maintaining reliability. Also, the approximately 5\% decrease in average computation time and the up to 60\% decrease in maximum time from \drafto-GN to \drafto validate the efficiency improvement, especially in the worst case, of the two-phase globalization policy. 

GPMP2 outperforms other optimization-based planners (CHOMP and TrajOpt) and strikes a good balance between efficiency and reliability. However, \drafto still runs $\times$2 to $\times$6 faster than GPMP2 and achieves a much higher success rate, 92\%--97\%, compared to GPMP2's 74\%--83\%. TrajOpt performs worst because it performs constrained optimization to step within the trust region, and its strict acceptance policy further constrains exploration.  

RRT-C achieves the best performance among all sampling-based planners and has a slightly higher success rate due to its probabilistic completeness. However, its average computation time is $\times$7 to $\times$120 of \drafto, and its trajectory is much more jerky, $\times$0.5 to $\times$3.5 rougher than \drafto. PRM follows RRT-C, and RRT* performs worst, but with a less jerky trajectory due to its tree-rewiring process.

\paragraph{Comparison across the scenes}

In \emph{cage}, the most difficult scene of the unconstrained tasks as shown in \figref{fig:simu_single_arm_01}, \drafto outperforms all the others in both success rate and computation time. It achieves a $\times$6 speedup over GPMP2 and a $\times$120 speedup over RRT-C while maintaining the highest reliability. 
Though RRT-C achieves a higher success rate in \emph{kitchen} and \emph{table\_under\_pick} by 7\% and 1\%, \drafto is $\times$24 and $\times$7 faster in these two scenes. 

Comparison between \tabref{tab:single_arm} and \tabref{tab:single_arm_constr} shows that the task constraints reduce the feasible space and that the additional computational cost arises. \drafto takes $\times$1.8 to $\times$2.5 computation resources in the constrained tasks over the ones in the unconstrained tasks on average. Its success rate decreases from 92\% in \emph{kitchen} and 97\% in \emph{table\_under\_pick} to 88\% and 95\%, respectively. Similarly, the success rates of CHOMP and RRT-C decrease from 85\%\,/\,36\% and 99\%\,/\,98\% to 10\%\,/\,9\% and 86\%\,/\,72\%, respectively. Meanwhile, their computation costs increase by around $\times$10.

\subsubsection{Dual-arm case}

\begin{figure}[!htp]
    \centering
    \hfill
    \begin{subfigure}[t]{1\linewidth}
        \centering
        \includegraphics[width=0.32\linewidth]{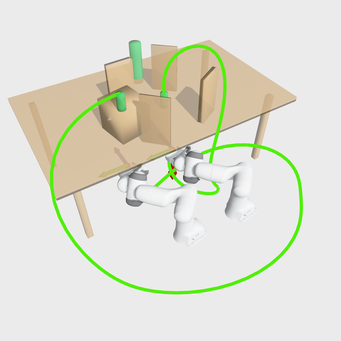}
        \includegraphics[width=0.32\linewidth]{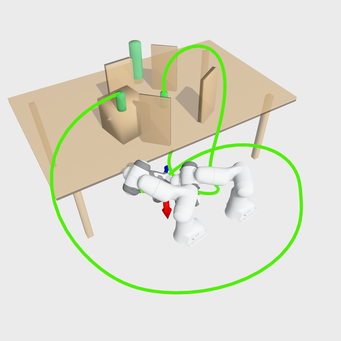}
        \includegraphics[width=0.32\linewidth]{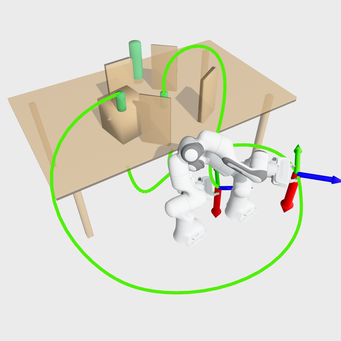}
        \\[2pt]
        \includegraphics[width=0.32\linewidth]{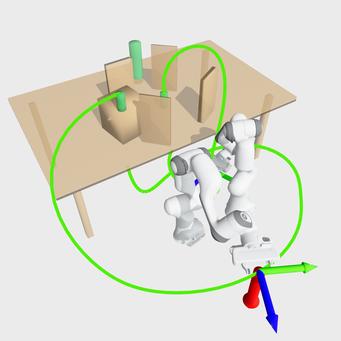}
        \includegraphics[width=0.32\linewidth]{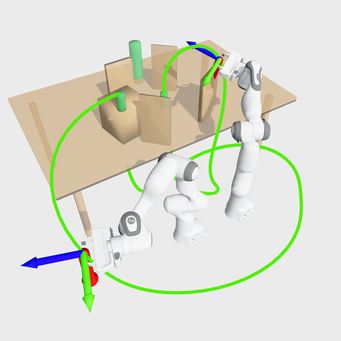}
        \includegraphics[width=0.32\linewidth]{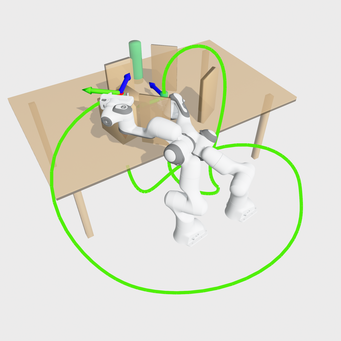}
    \end{subfigure}
    \hfill
    \caption{Simulation results of the dual FR3 arms in \emph{table\_constr}: the end-effector $x$-axis (red arrow) is constrained to align with the world $z$-axis; representative snapshots along the trajectory are shown (green). }
    \label{fig:sim_dual_arm}
\end{figure}

As shown in \figref{fig:sim_dual_arm}, feasible initial and goal waypoints not only need to meet the safety requirement, including the obstacle-, self-, and cross-collision checks, but also should satisfy the task constraint. Therefore, only 40 feasible tasks are generated for the benchmark tests of both the unconstrained and constrained tasks. The remainder of this section provides a detailed analysis of the test results. 

\begin{table*}[!htp]
\centering
\setlength{\tabcolsep}{2.4pt}
\renewcommand{\arraystretch}{0.95}
\caption{Benchmark results across dual-arm planning scenes.}
\label{tab:dual_arm}
\begin{tabular}{@{}llccccccccc@{}}
\toprule
Scene & Metrics & \drafto-GN & \drafto & FACTO & CHOMP & TrajOpt & GPMP2 & RRT-C & RRT* & PRM \\
\midrule

\multirow{3}{*}{table}
& S (\%)      & 85.0 & 85.0 & 92.5 & 27.5 & 35.0 & 47.5 & 90.0 & 12.5 & 12.5 \\
& T (avg/max) & 0.102\,/\,0.303 & 0.081\,/\,0.201 & 0.242\,/\,0.853 & 0.438\,/\,0.501 & 1.250\,/\,1.517 & 0.312\,/\,0.822 & 2.027\,/\,8.391 & 7.100\,/\,8.162 & 5.339\,/\,10.006 \\
& R (avg/max) & 237.1\,/\,363.8 & 236.1\,/\,363.8 & 250.6\,/\,466.4 & 88.9\,/\,108.3 & 125.1\,/\,177.7 & 60.0\,/\,88.6 & 512.0\,/\,1013.2 & 337.1\,/\,628.8 & 498.9\,/\,847.3 \\
\midrule

\multirow{3}{*}{table\_constr}
& S (\%)      & 87.5 & 87.5 & 87.5 & 7.5 & - & - & 47.5 & - & - \\
& T (avg/max) & 0.113\,/\,0.462 & 0.099\,/\,0.307 & 0.370\,/\,0.656 & 3.533\,/\,4.211 & - & - & 4.232\,/\,8.609 & - & - \\
& R (avg/max) & 295.8\,/\,399.7 & 295.6\,/\,399.7 & 322.2\,/\,488.1 & 95.6\,/\,140.2 & - & - & 1122.3\,/\,2674.4 & - & - \\
\bottomrule
\end{tabular}
\end{table*}

\paragraph{Comparison across the algorithms}

\tabref{tab:dual_arm} shows FACTO attains highest success rate and \drafto achieves the lowest computation cost. Similar to the single-arm case, \drafto is $\times$2.5 faster than FACTO and consumes 20\% less computation cost than that of \drafto-GN. GPMP2 remains the second-highest success rate among optimization-based planners and the second-highest computational efficiency among all planners. TrajOpt and CHOMP show the lowest efficiency and reliability among the optimization-based planners, respectively. RRT-C achieves the second-best success rate and the shortest solving time among sampling-based planners, whereas RRT* and PRM exhibit the worst performance in both efficiency and reliability. 

\paragraph{Comparison between the unconstrained and the constrained tasks}

The comparison between \emph{table\_under\_pick} and \emph{table\_under\_pick\_constr} in the dual-arm case aligns with the corresponding comparison in the single-arm case. The success rate of all the planners, except for \drafto and \drafto-GN, decreases, while their computation time increases. As for \drafto and \drafto-GN, the changes in efficiency and reliability are minor, with only a side effect on their roughness when moving from the unconstrained to the constrained task.

\subsection{Real-world Experiment}
\label{sec:hardware}

\subsubsection{Set up}

This section presents the results of the single-arm experiment, which is conducted on FR3 via the Franka control interface (FCI) on a NUC running Ubuntu 22.04.

\subsubsection{Planning-execution workflow}

Given the information of work environment, robot kinematics/geometry, initial/goal waypoints, and task requirements, \drafto first performs offline planning to generate a feasible and safe motion parameterized by $\bm\psi$. After the joint-feasibility and collision checks, \drafto adopts \texttt{facto::TimeScaler} to estimate the execution time $T$. The optimized trajectory is then published and executed via \texttt{libfranka} using UDP in an RT-kernel. 

\begin{figure}[!htbp]
    \centering
    \hfill
    \begin{subfigure}[t]{1\linewidth}
        \centering
        \includegraphics[width=0.32\linewidth]{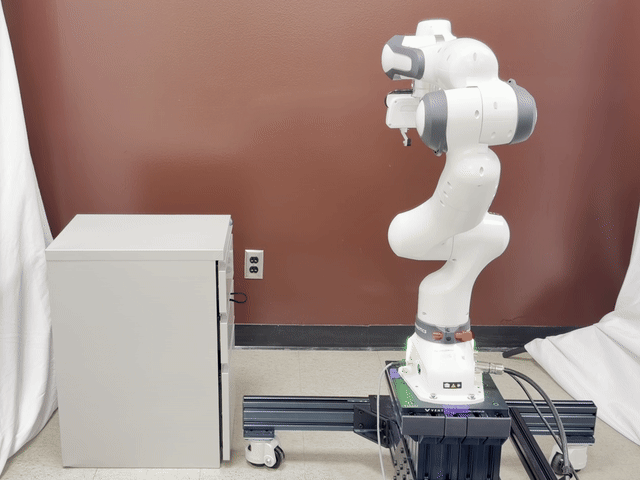}
        \includegraphics[width=0.32\linewidth]{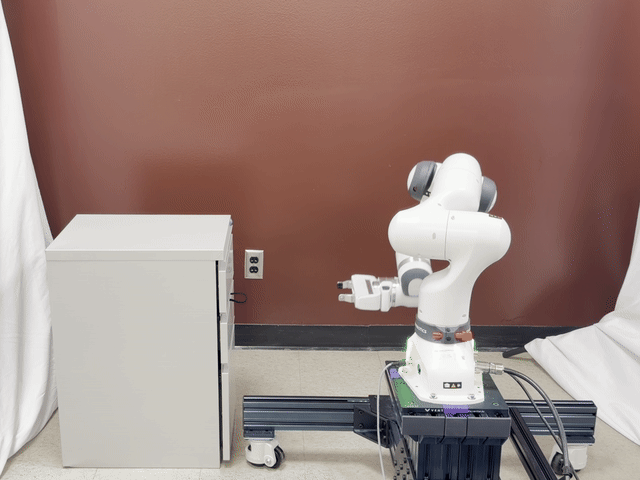}
        \includegraphics[width=0.32\linewidth]{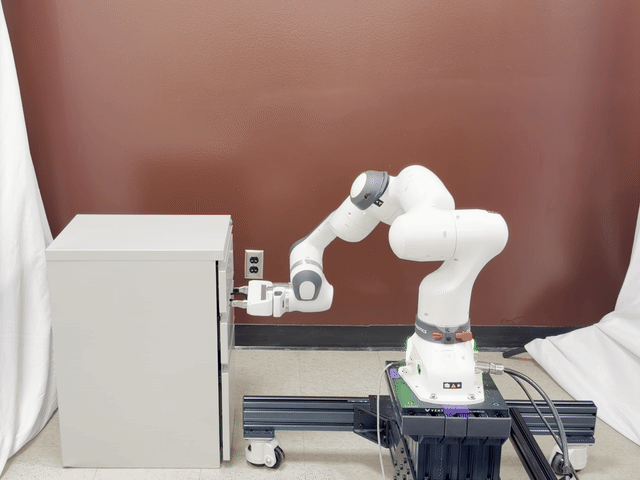}
        \caption{FR3 moves from an initial configuration to grab a drawer. }
    \end{subfigure}
    \hfill
    \\[2pt]
    \hfill
    \begin{subfigure}[t]{1\linewidth}
        \centering
        \includegraphics[width=0.32\linewidth]{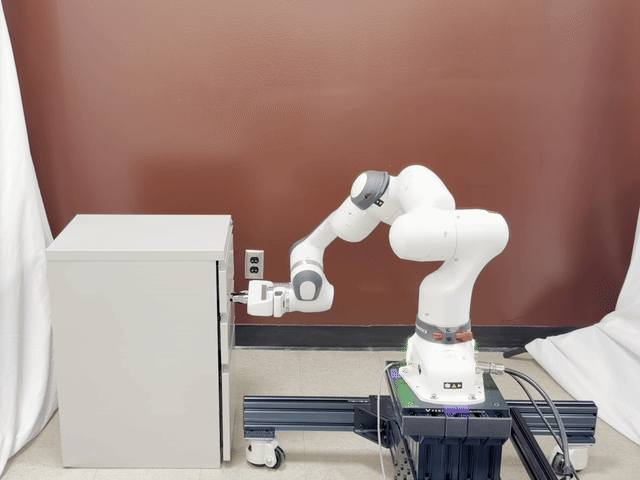}
        \includegraphics[width=0.32\linewidth]{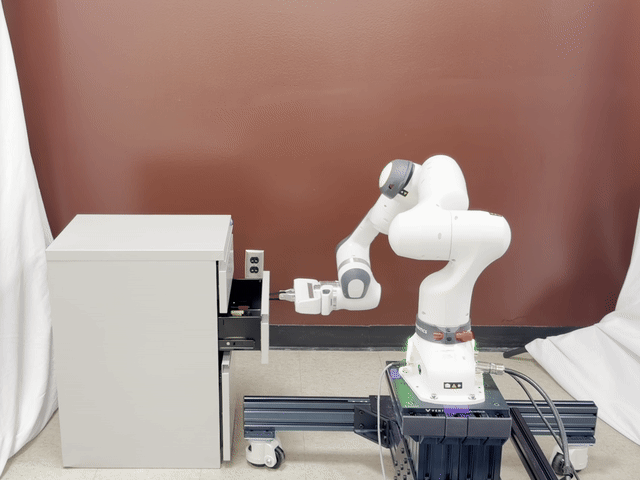}
        \includegraphics[width=0.32\linewidth]{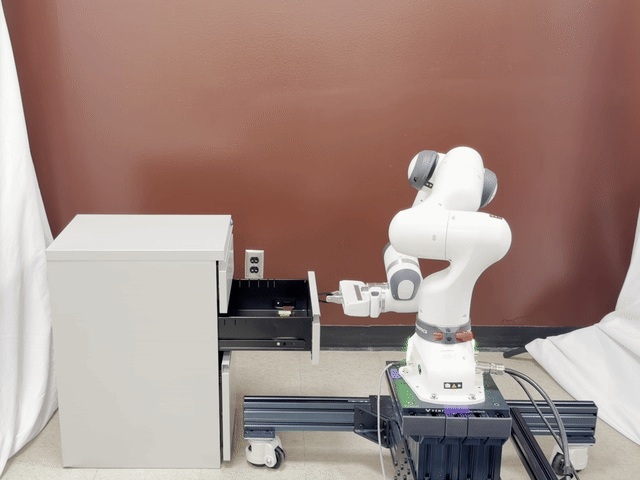}
        \caption{FR3 pulls the drawer out of the cabinet along a straight line. }
    \end{subfigure}
    \hfill
    \\[2pt]
    \hfill
    \begin{subfigure}[t]{1\linewidth}
        \centering
        \includegraphics[width=0.32\linewidth]{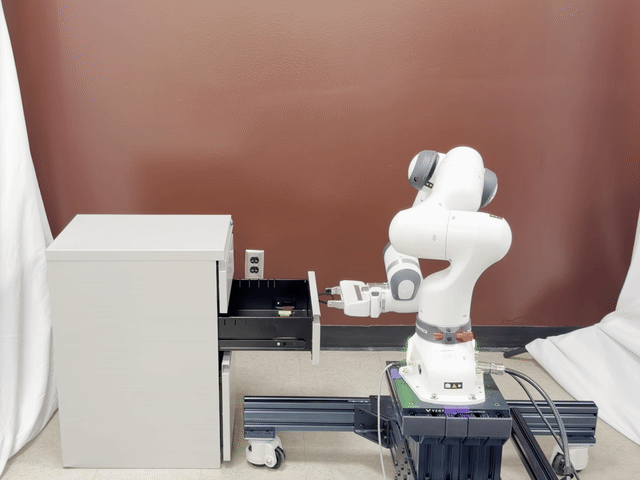}
        \includegraphics[width=0.32\linewidth]{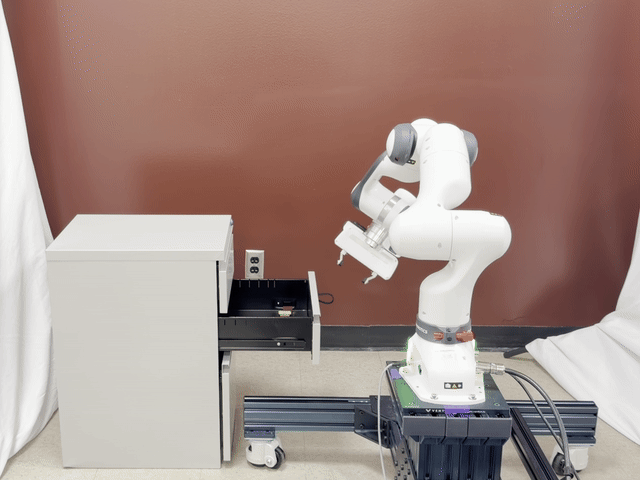}
        \includegraphics[width=0.32\linewidth]{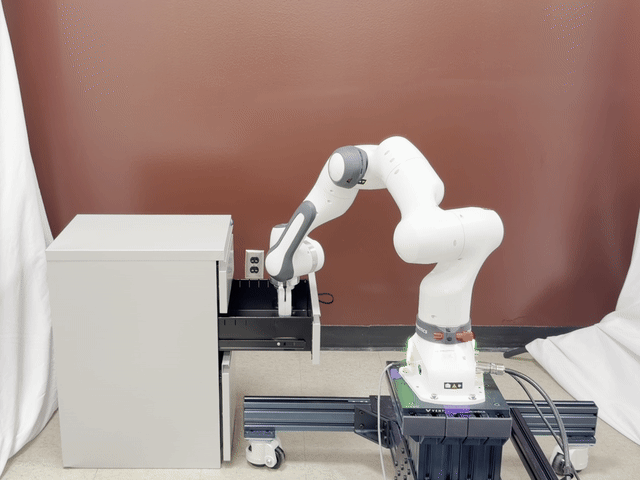}
        \caption{FR3 moves to grab an object inside the drawer. }
    \end{subfigure}
    \hfill
    \\[2pt]
    \hfill
    \begin{subfigure}[t]{1\linewidth}
        \centering
        \includegraphics[width=0.32\linewidth]{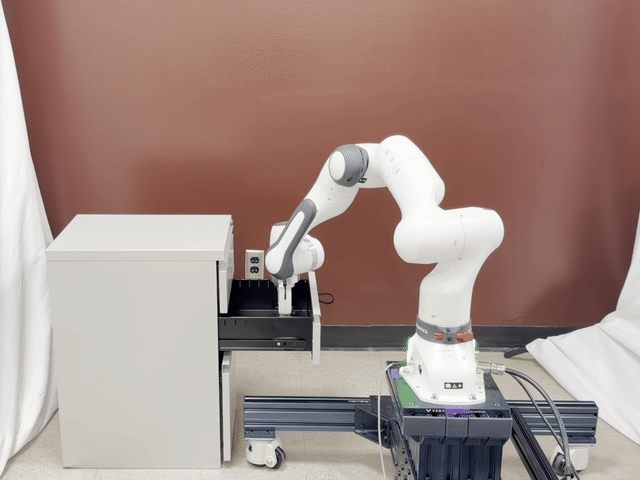}
        \includegraphics[width=0.32\linewidth]{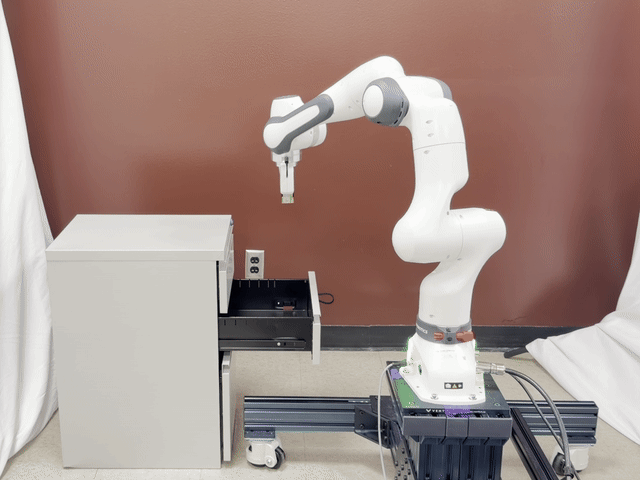}
        \includegraphics[width=0.32\linewidth]{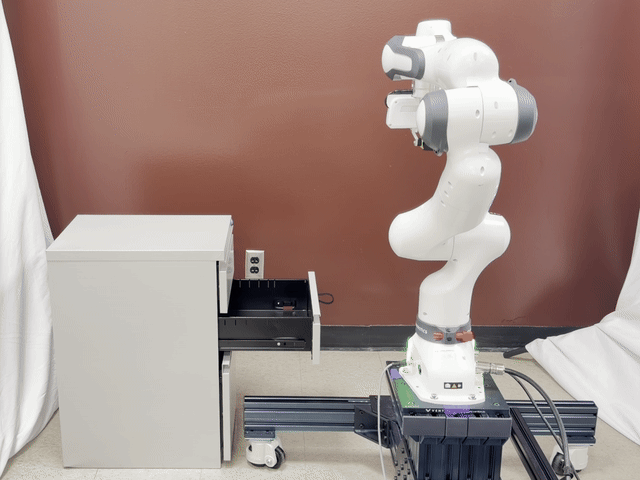}
        \caption{FR3 unplugs the object vertically and returns to the initial state. }
    \end{subfigure}
    \hfill
    \caption{Physical experiment of an FR3 arm grabbing an object inside a drawer. }
    \label{fig:exp}
\end{figure}

\figref{fig:exp} shows how the trajectory, parameterized by basis functions along with their coefficients, is executed smoothly and efficiently by FR3 to pick an object inside a drawer.

\section{Conclusion and Discussion}
\label{sec:conclusion}

This work introduces a new algorithm, \drafto, for fast and reliable trajectory optimization that satisfies smoothness, safety, joint limits, and task requirements. Its decoupled framework, which combines the reduced GN and the constrained QP under the two-phase globalization rule, improves computational efficiency while preserving feasibility. Its convergence is theoretically discussed, and its efficiency and reliability are validated through over 1000 benchmark tests on both single- and dual-arm unconstrained/constrained tasks. The physical experiment further validates the implementation of the planning-execution workflows on an FR3 robot.

\bibliographystyle{IEEEtran}
\bibliography{reference}

\appendix

\begin{table}[!htp]
\centering
\setlength{\tabcolsep}{2pt}        
\renewcommand{\arraystretch}{0.95} 
\caption{Benchmark results across single-arm planning scenes.}
\label{tab:benchmark_appendix_scenes}

\resizebox{\columnwidth}{!}{%
\begin{tabular}{@{}llccccccccc@{}} 
\toprule
Scene & Metrics & \drafto-GN & \drafto & FACTO & CHOMP & TrajOpt & GPMP2 & RRT-C & RRT* & PRM \\
\midrule

\multirow{5}{*}{bookshelf\_small}
& S (\%)      & 96 & 96 & 98 & 98 & 67 & 97 & 96 & 47 & 78 \\
& T avg       & 0.0097 & 0.0092 & 0.0387 & 0.1010 & 0.1297 & 0.0624 & 0.6753 & 1.5649 & 0.8481 \\
& T max       & 0.0954 & 0.0738 & 0.1235 & 0.2410 & 0.4280 & 0.1594 & 9.5129 & 9.9939 & 8.4673 \\
& R avg  & 50.12 & 50.07 & 52.02 & 36.59 & 96.17 & 22.52 & 75.15 & 61.16 & 178.60 \\
& R max  & 102.18 & 97.22 & 85.62 & 54.06 & 579.64 & 37.42 & 555.34 & 198.49 & 408.42 \\
\midrule

\multirow{5}{*}{bookshelf\_tall}
& S (\%)      & 98 & 98 & 98 & 98 & 84 & 97 & 93 & 28 & 83 \\
& T avg       & 0.0075 & 0.0077 & 0.0272 & 0.0886 & 0.1348 & 0.0631 & 0.4828 & 1.4760 & 1.5758 \\
& T max       & 0.0220 & 0.0209 & 0.0785 & 0.2347 & 0.3479 & 0.2592 & 8.0935 & 9.3306 & 9.2312 \\
& R avg  & 49.63 & 49.63 & 40.97 & 37.47 & 61.78 & 22.52 & 79.65 & 56.34 & 116.58 \\
& R max  & 68.17 & 68.17 & 85.75 & 59.75 & 235.97 & 37.42 & 455.57 & 172.87 & 553.57 \\
\midrule

\multirow{5}{*}{bookshelf\_thin}
& S (\%)      & 100 & 100 & 100 & 100 & 63 & 94 & 97 & 25 & 61 \\
& T avg       & 0.0104 & 0.0101 & 0.0478 & 0.0926 & 0.1426 & 0.0666 & 0.8085 & 3.2522 & 2.2643 \\
& T max       & 0.0250 & 0.0218 & 0.1043 & 0.2170 & 0.4900 & 0.1422 & 7.6820 & 9.6943 & 10.3143 \\
& R avg  & 47.60 & 47.60 & 39.11 & 34.67 & 65.64 & 21.93 & 92.78 & 65.60 & 116.88 \\
& R max  & 70.47 & 70.47 & 76.45 & 50.60 & 297.89 & 37.95 & 532.78 & 169.41 & 536.14 \\
\midrule

\multirow{5}{*}{box}
& S (\%)      & 96 & 96 & 98 & 98 & 22 & 93 & 99 & 4 & 85 \\
& T avg       & 0.0187 & 0.0164 & 0.0546 & 0.0961 & 0.2428 & 0.0590 & 0.2927 & 4.7499 & 1.9173 \\
& T max       & 0.1136 & 0.0643 & 0.2863 & 0.2390 & 0.4431 & 0.3306 & 4.4057 & 9.4519 & 9.5398 \\
& R avg  & 49.74 & 43.53 & 53.14 & 33.94 & 225.98 & 22.48 & 120.63 & 68.01 & 178.07 \\
& R max  & 98.64 & 87.03 & 103.93 & 46.99 & 472.49 & 40.59 & 466.64 & 139.62 & 443.24 \\
\midrule

\multirow{5}{*}{table\_pick}
& S (\%)      & 97 & 97 & 99 & 87 & 56 & 97 & 98 & 33 & 85 \\
& T avg       & 0.0104 & 0.0103 & 0.0421 & 0.1310 & 0.1517 & 0.0264 & 0.2306 & 2.9695 & 1.3845 \\
& T max       & 0.0243 & 0.0219 & 0.1034 & 0.3211 & 0.4177 & 0.2437 & 4.4521 & 9.8163 & 7.4833 \\
& R avg  & 55.01 & 55.01 & 55.64 & 37.56 & 93.60 & 22.42 & 65.06 & 46.74 & 148.34 \\
& R max  & 100.41 & 100.41 & 106.45 & 53.73 & 285.39 & 32.81 & 345.16 & 131.10 & 360.64 \\
\bottomrule
\end{tabular}%
}
\end{table}

\end{document}